\pdfoutput=1
\documentclass[11pt]{article}

\usepackage{acl}

\usepackage{times}
\usepackage{latexsym}
\usepackage{algorithm}
\usepackage{algorithmic}
\usepackage{amsmath}
\usepackage{graphicx} 
\usepackage{pifont}
\usepackage{latexsym}
\usepackage{booktabs}
\usepackage{graphicx}
\usepackage{times}
\usepackage{latexsym}
\usepackage{lineno}
\usepackage{caption}
\usepackage{subcaption}
\usepackage{colortbl}
\usepackage{courier}
\usepackage[T1]{fontenc}
\usepackage{amsmath}
\usepackage{amssymb}
\usepackage{multirow}
\usepackage{dsfont}
\usepackage{url}
\usepackage{makecell}
\usepackage{xcolor}
\usepackage{bm}
\usepackage{tikz}
\usepackage{url}
\usepackage{relsize}
\usepackage{hyperref}
\usepackage{enumitem}
\usepackage{makecell}

\usepackage[T1]{fontenc}
\usepackage[utf8]{inputenc}

\usepackage{microtype}

\newcommand\blfootnote[1]{%
  \begingroup
  \renewcommand\thefootnote{}\footnote{#1}%
  \addtocounter{footnote}{-1}%
  \endgroup
}

\title{\textsc{X-Turing}: Towards an Enhanced and Efficient Turing Test \\for Long-Term Dialogue Agents}

\author{
Weiqi Wu,
Hongqiu Wu,
Hai Zhao$^\dag$ \\
School of Computer Science, Shanghai Jiao Tong University \\
\texttt{\{wuwq1022,wuhongqiu\}@sjtu.edu.cn,}\\
\texttt{zhaohai@cs.sjtu.edu.cn}}

\begin{document}
\maketitle

\blfootnote{$^\dag$ Corresponding author. This research was supported by the Joint Research Project of Yangtze River Delta Science and Technology Innovation Community (No. 2022CSJGG1400).}
\blfootnote{$^\ddag$ Code is available at \url{https://github.com/vickywu1022/X-Turing}.}

\begin{abstract}

The Turing test examines whether AIs exhibit human-like behaviour in natural language conversations. The traditional setting 
% adopts rigid turn-based dialogues, 
limits each participant to one message at a time and requires constant human participation. This fails to reflect a natural conversational style and hinders the evaluation of dialogue agents based on Large Language Models (LLMs) in complex and prolonged interactions.
This paper proposes \textbf{\textsc{X-Turing}}, which enhances the original test with a \textit{burst dialogue} pattern, allowing more dynamic exchanges using consecutive messages. It further reduces human workload by iteratively generating dialogues that simulate the long-term interaction between the agent and a human to compose the majority of the test process. With the \textit{pseudo-dialogue} history, the agent then engages in a shorter dialogue with a real human, which is paired with a human-human conversation on the same topic to be judged using questionnaires. 
We introduce the \textit{X-Turn Pass-Rate} metric to assess the human likeness of LLMs across varying durations. While LLMs like GPT-4 initially perform well, achieving pass rates of 51.9\% and 38.9\% during 3 turns and 10 turns of dialogues respectively, their performance drops as the dialogue progresses, which underscores the difficulty in maintaining consistency in the long term.$^\ddag$ 

\end{abstract}

\section{Introduction}

The Turing Test, originally called the imitation game by Alan Turing in 1950, is a test of whether a machine can behave indistinguishably from a human \cite{Turing1950ComputingMA}. During the test, a human integrator converses in natural language with both a human and an AI, then tries to identify which is which. The AI is considered to pass the test if it can convincingly mimic human responses so that the integrator misidentifies it as a human participant. Contemporary Large Language Models (LLMs) \cite{DBLP:journals/corr/abs-2303-08774,DBLP:journals/corr/abs-2307-09288,DBLP:journals/corr/abs-2310-06825,DBLP:journals/corr/abs-2309-10305,qwen,TheC3} demonstrate remarkable language generation capabilities in various scenarios \cite{wu2024scalingvirtualworlddeltaengine}. These advancements have sparked increasing interest in whether the latest LLMs can pass the Turing test \cite{Jones2024PeopleCD}. 

However, the traditional Turing test has several limitations: (1) The dialogue follows a strict ping-pong structure, unlike natural communication where people may send multiple messages in a row to convey a statement. (2) The need for continuous human involvement in interactions with the test subject limits test duration to under ten minutes, leaving long-term evaluation unexplored. (3) Measuring test duration in minutes ignores individual differences in reading, thinking, and typing speeds, potentially disadvantaging some participants.

\begin{figure*}[ht]
    \centering
    \includegraphics[width=0.9\linewidth]{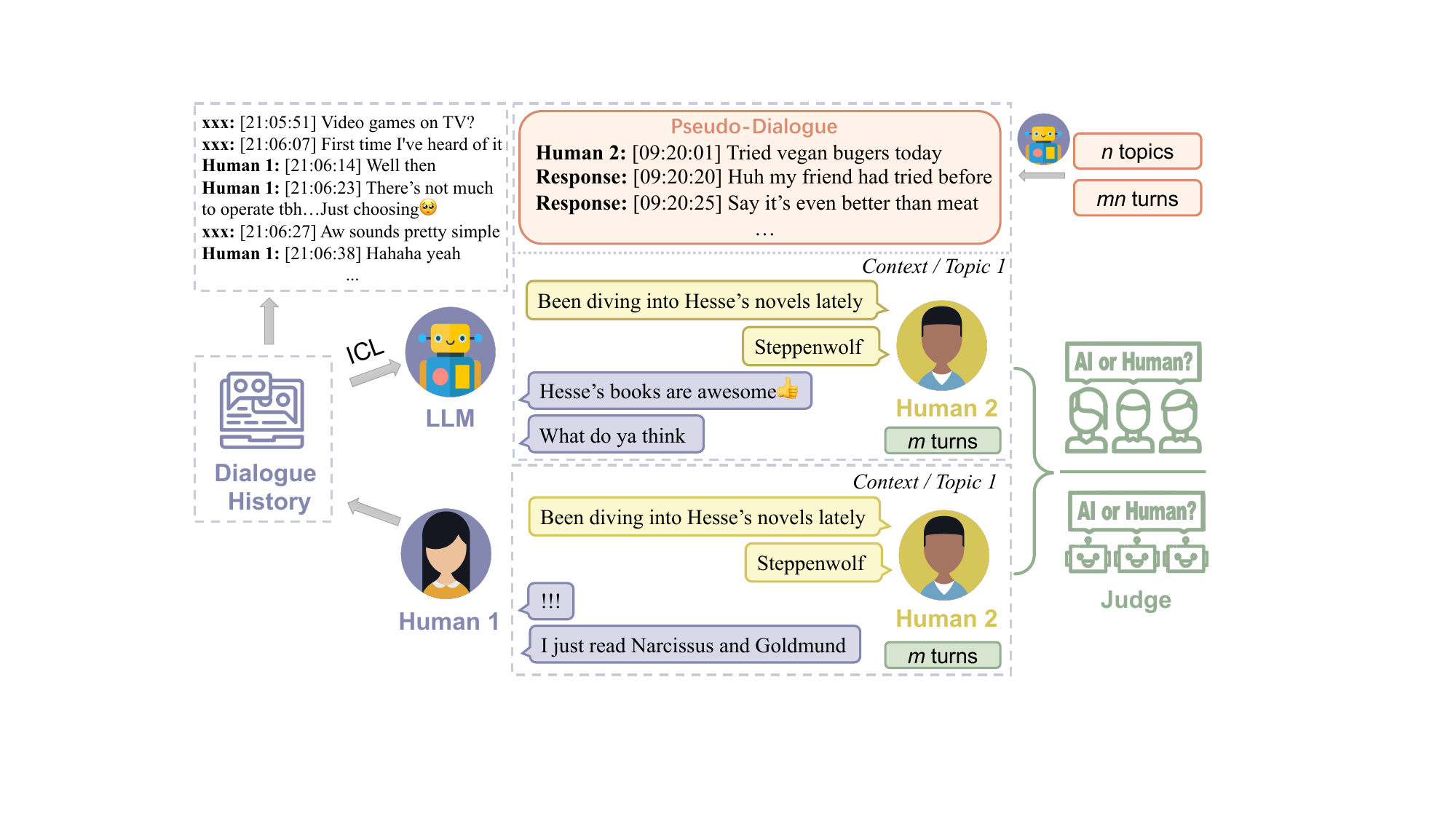}
    \caption{Overview of the \textbf{\textsc{X-Turing}}. Prompted by the dialogue history of a target person (Human 1), the LLM chats with another human (Human 2) after a specific turn of pseudo-dialogue to simulate its long-term interaction performance. The judges then distinguish between the LLM and Human 1 when they each converse with Human 2.}
    \label{fig:intro}
\end{figure*}
 
To address these issues, we propose \textbf{\textsc{X-Turing}}, a new framework designed to conduct an efficient Turing test for LLM-based dialogue agents with an enhanced dialogue pattern that aligns with daily communication among humans. Firstly, we break the constraint of a single message each turn to allow users to send multiple messages in rapid succession without waiting for a response. As illustrated in Figure \ref{fig:intro}, this dynamic exchange mirrors natural text-based human interactions and challenges LLMs to maintain coherence and relevance in more diverse interactions. We construct chatbots for the test by prompting LLMs with social platform dialogue records from a real-life human.

Given the exhausting nature of long-term Turing test, we propose enhancing its efficiency by approximating real human-machine interactions through dialogue generation techniques \cite{Soudani2024ASO}, and using questionnaires instead of having evaluators engage in extended conversations. According to psychological research on the serial position effect \cite{Murdock1962TheSP}, the first and last parts of a sequence are likely to be most memorable and thus critical for judgment. As it is more challenging for LLMs to maintain consistency and coherence over long-term interactions \cite{Park2023GenerativeAI,Bubeck2023SparksOA}, our evaluation focuses on the last few turns of the conversation. 
% Instead of requiring human involvement during the previous turns of real-time interaction, 
Therefore, we propose pseudo-dialogue generation to allow the model to direct this progression on its own. It involves iteratively generating multi-turn dialogues to reflect the LLM's evolving performance over time. Human participants only need to converse subsequently for a few more turns, as depicted in Figure \ref{fig:intro}. By focusing the evaluation on the final interaction turns, we can provide a comprehensive assessment while significantly reducing the human workload, effectively making the Turing test scalable in terms of interaction length.

Our contributions can be summarized as follows:
\begin{itemize}
    \item We extend the classical Turing test with burst dialogue, more closely reflecting natural human communication.
    \item We employ dialogue generation techniques to automatically simulate user-machine interactions, making longer Turing tests feasible.
    \item We introduce a new metric, \textit{X-Turn Pass-Rate}, to fairly assess the ability of LLMs to maintain human-like over a set of interaction turns.
    \item We perform extensive experiments using the \textbf{\textsc{X-Turing}}  framework to evaluate the latest LLMs and explore key factors influencing test outcomes.
\end{itemize}

\section{Related Work}

\paragraph{Turing Test}
Turing test and its variations have been extensively applied to assess AI 
%in exhibiting human-like abilities 
across multiple domains, including dialogue interaction \cite{Jones2024PeopleCD,Lowe2017TowardsAA}, question answering\cite{Ng2024HowWC}, text generation \cite{Helm2023AST,Uchendu2021TURINGBENCHAB}, and visual tasks \cite{imanpour2024visualcounterturingtest,Zhang2022HumanOM}. These evaluations, which involve distinguishing between human-generated and AI-generated content \cite{Gao2023TheTQ} or inferring from solely the content \cite{Jannai2023HumanON}, serve as essential benchmarks for AI models. Existing works proposing the limitations of Turing tests focus on whether passing the test indicates intelligence accurately \cite{DeFreitas2020TowardsAH} or comprehensively \cite{VideoTuringTest2023}. We rather focus on its feasibility issue in long-term evaluation and enhance the flexibility of the communication pattern.

% Apart from extending the test to a more complex dialogue setting, we further suggest an automated approach to reduce the significant human effort required by Turing tests, being the first to address the feasibility issue.

\paragraph{Role-play with LLMs}
A common tactic in Turing tests is to assign LLMs the role of a specific human character \cite{Jannai2023HumanON}, leveraging their proficiency in adopting varied attributes and conversational styles \cite{wu-etal-2024-role,Shanahan2023RolePW,wang2023does}. Prompting the model with character profiles \cite{li2023chatharuhi,tao2023rolecraftglm,Wang2023RoleLLMBE,chen-etal-2023-large} or key utterances \cite{han-etal-2022-meet,Boyd2020LargeSM} proves to be a more efficient method, compared with fine-tuning on massive character data \cite{shao-etal-2023-character,Zhou2023CharacterGLMCC,lu2024large}. By utilizing the dialogue history of the designated individual, we can develop a role-playing chatbot that effectively handles various dialogue forms and scenarios.

\paragraph{Dialogue Generation}

The development of synthetic dialogue data has provided a more efficient and scalable approach to data creation and augmentation \cite{Soudani2024ASO,Chen2024RecentTI,Zhou2017EmotionalCM}. Advancements in dialogue generation have been facilitated by LLMs \cite{Abdullin2024SyntheticDD,Chen2023ControllableMD,Chen2023PLACESPL,Kim2022SODAMD,Abbasiantaeb2023LetTL}, which excel at creating contextually relevant and diverse conversations by prompted turn-by-turn \cite{Lee2022PERSONACHATGENGP,Kim2022BotsTalkMF} or even in just one go \cite{Jandaghi2023FaithfulPC,Chen2023PLACESPL}. By iteratively generating dialogues that imitate interactions with a human, we can efficiently simulate the Turing test over extensive turns and facilitate further judgment.

\section{\textbf{\textsc{X-Turing}}}

\subsection{Burst Dialogue v.s. Ping-pong Dialogue}

Conversations with a ping-pong structure are considered the standard form of human-machine interaction and are commonly used in the Turing Test, where each user message is followed by a system response to form a strict back-and-forth pattern. However, in real life, people usually send multiple messages in rapid succession without waiting for the other's response and the response can also consist of multiple messages. We define this communication pattern as \textit{Burst Dialogue} and delve into the performance of LLMs in this setting.

In a ping-pong dialogue, a \textit{turn} is naturally formed by pairing each user message with the corresponding system response. In contrast, burst dialogue involves a more flexible information exchange, where interruption and continuation can happen frequently for both sides. For simplicity, a turn of burst dialogue is defined as a consecutive series of user messages followed by the system's responses before the next user message. Despite potential unbalanced exchange, burst dialogue is more dynamic and aligns more closely with the natural human communication pattern. It challenges the model to handle overlapping and rapidly evolving conversational contexts, thereby testing its ability to maintain coherence and relevance in real-time interactions. Hence, we extend the classical ping-pong dialogue setting to assess how well LLMs manage more realistic conversational dynamics.

\subsection{Chatbot Construction}

The Turing test necessitates the deployment of a chatbot to judge its text-based conversation ability. 
Instead of manually crafting detailed profiles for individuals, we construct the dialogue agent based on real-life dialogue history, which provides essential clues for generating responses that emulate the target personality, linguistic style and content. To achieve this, we gather multi-turn text-based social platform chat records for each target individual, which are burst conversational data consisting of timestamps and message content. The LLM is then instructed to maintain consistency with the conversation style with the prompt presented in Appendix \ref{sec: role prompt}. 
For ping-pong dialogues, we retain only the first message of a consecutive series from the same person and manually filter out meaningless turns.

\begin{figure}[t]
    \centering
    \includegraphics[width=\linewidth]{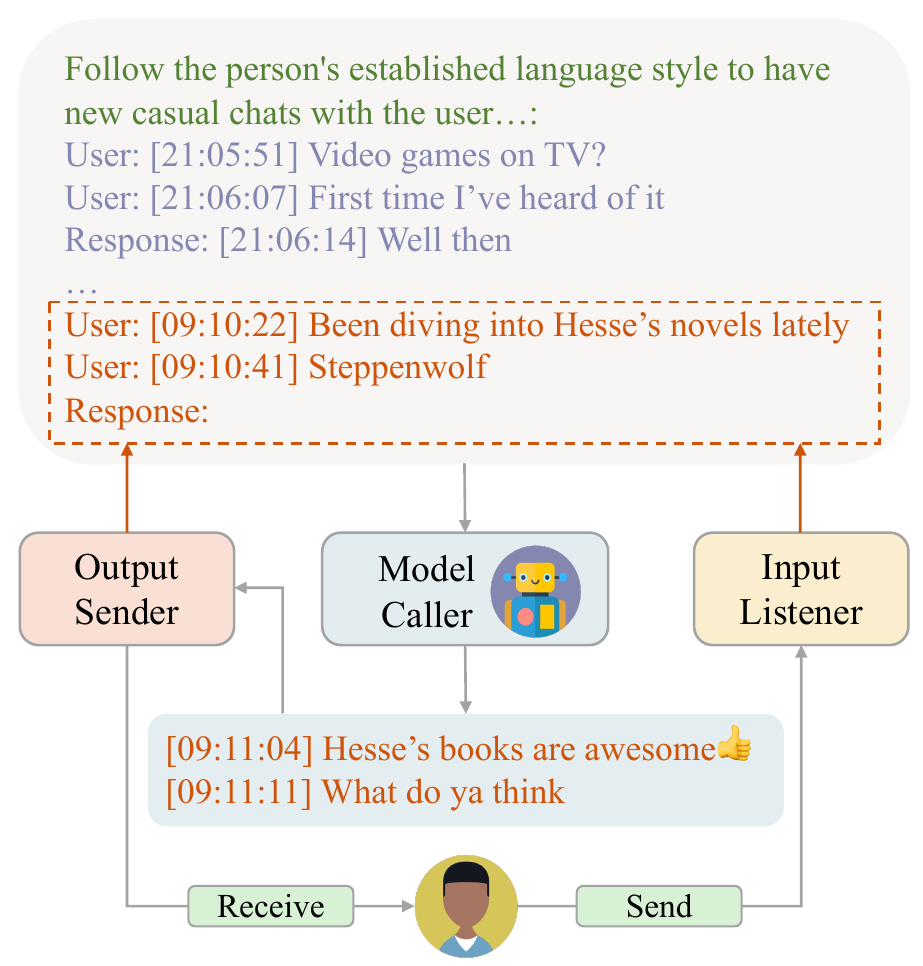}
    \caption{Chatbot system enabling burst dialogue.}
    \label{fig:chatbot}
\end{figure}

To implement a burst messaging chatbot, we employ three modules working synchronously as shown in Figure \ref{fig:chatbot}: 
$\bullet$  \textit{Input Listener}: to collect user messages; 
$\bullet$ \textit{Model Caller}: to query the LLM to generate responses; $\bullet$ \textit{Output Sender}: to manage the delivery of system responses based on the timestamps. The system introduces a brief time interval $t_1$ before processing the first batch of user inputs to allow the user to fully express themselves. After $t_1$, the Input Listener processes all user messages collectively for the Model Caller to generate a response. 

Following the given burst dialogue data and the instruction, the chatbot generates consecutive messages in a response that can be sent to the user and added to the dialogue history by the Output Sender based on their respective timestamps. In case of any timestamp errors, e.g., being earlier than the current timestamp, we resample the message sending time. A message consisting of $n$ characters is sent after a delay $d$ of around 0.3s per character to mimic the reading, thinking and typing speed of humans\cite{Jones2024PeopleCD}:
\begin{equation*}
    d = \mathcal{N}(0.3, 0.03) \times n
\end{equation*}
The Model Caller typically requires time to generate responses with LLMs. During this period, the Input Listener and Output Sender continuously work to maintain the dialogue history. Once a query is completed, if the user has sent new messages during this time, the Model Caller initiates the next query; otherwise, it waits for a time interval $t_1$ before processing the accumulated dialogue history to generate new responses. If new messages are generated, any unsent messages from the previous batch are cleared to maintain the conversation flow.

\begin{algorithm}[t]
\caption{Pseudo-Dialogue Generation}
\label{alg:pseudo_dialogue_generation}
\begin{algorithmic}[1]
\REQUIRE Number of Turns $m$, Dialogue History $H$, Topics $T=\{t_1, t_2, \dots, t_n\}$
\ENSURE Pseudo-Dialogue $D$
\STATE $D \gets \{\}$
\FOR{each $t_i \in T$}
    \STATE $pd \gets \{\}$
    \WHILE{\text{len}($pd$) < $m$}
        \STATE $pd \gets pd \cup \text{LLM}(t_i, m, H)$
        \IF{\text{len}($pd$) > $m$}
            \STATE $pd \gets pd[:$$m$$]$
        \ENDIF
    \ENDWHILE
    \STATE $D \gets D \cup pd$
    \STATE $H \gets concat(H, str(pd)) $
\ENDFOR
\RETURN $D$
\end{algorithmic}
\end{algorithm}

\subsection{Pseudo-Dialogue Generation}

To enhance the efficiency of the Turing test, we propose to let the LLMs direct the progression of the test by synthesizing extended user-machine interaction through dialogue generation. Classical open-domain conversational data generation generally includes three steps: (1) Input generation: Provide fundamental information to initiate the conversation that specifies the topic and personalized style, i.e., conversation seed; (2) Utterance generation: Utilize the conversation seed to generate a multi-turn conversation; (3) Quality filtering: Eliminate samples that fail to meet the quality criteria \cite{Soudani2024ASO}. The quality filtering step is omitted as our dialogue generation process aims to reflect the genuine evolving performance of the LLM over prolonged interaction.
\begin{figure}[htbp]
    \centering
    \includegraphics[width=\linewidth]{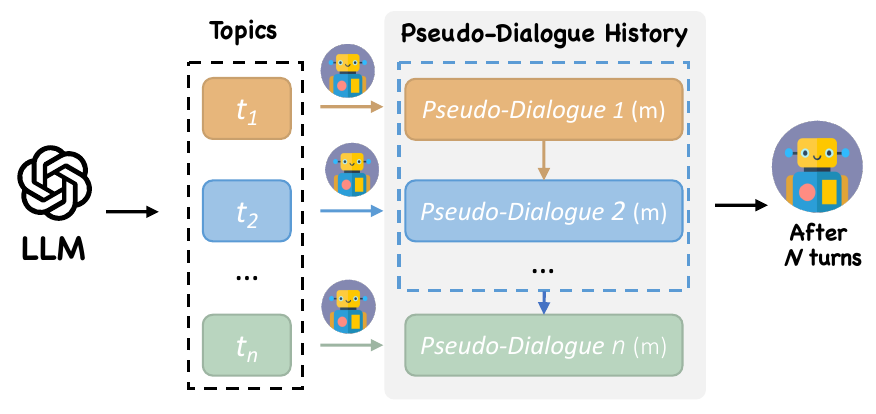}
    \caption{Pipeline of Pseudo-Dialogue Generation.}
    \label{fig:pdg}
\end{figure}

To generate $N = mn$ turns of pseudo-dialogue, we first utilize the LLM to produce $n$ distinct topics $T=\{t_1, t_2, \dots, t_n\}$. Based on the dialogue history $H$ of the target individual, we iteratively prompt the LLM to generate an $m$-turn dialogue between the individual and another human for each topic. If a dialogue does not have enough turns, we continue generating dialogue for the current topic until the requirement is met. If the dialogue exceeds $m$ turns, the excess turns are truncated. The generated dialogue is appended to the dialogue history $H$ for subsequent iteration, as detailed in Algorithm \ref{alg:pseudo_dialogue_generation}. The prompts used to generate topics and pseudo-dialogue are demonstrated in \ref{sec: pdg prompt}.

The $N$-turn pseudo-dialogues are incorporated into the prompt of the chatbot, which then interacts with a human for an additional $m_T$ turns to evaluate the performance of LLMs over $(N+m_T)$ turns of interaction by judging the last $m_T$-turn dialogue. % Without loss of generality, we set $m_T$ to 10.

\begin{figure}[t]
    \centering
    \includegraphics[width=0.95\linewidth]{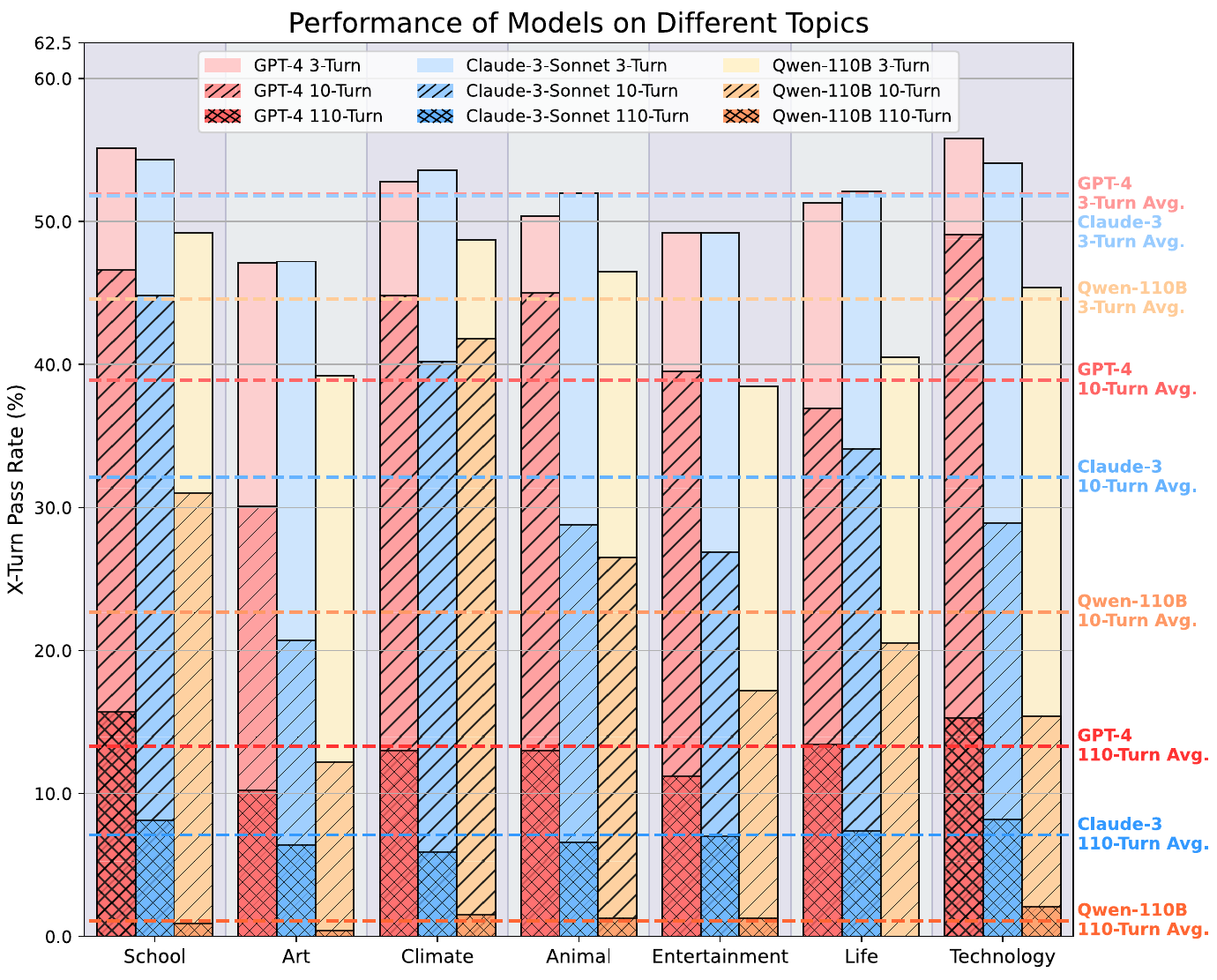}
    \caption{Topics covered in the test.}
    \label{fig:topic}
\end{figure}

\section{Experiments}

\begin{table*}[t]
\centering
\small
\setlength{\tabcolsep}{1.5mm}{
\begin{tabular}{cccc|ccc|ccc|ccc}
\toprule
\multirow{2}{*}{\textbf{Model}} &
  \multirow{2}{*}{\textbf{Duration}} &
  \multicolumn{2}{c|}{\textbf{\# Words}} &
  \multicolumn{3}{c|}{\textbf{Human-Judge}} &
  \multicolumn{3}{c|}{\textbf{GPT4-Judge}} &
  \multicolumn{3}{c}{\textbf{Qwen-Judge}} \\
                                    &          & P-P & Burst & P-P  & Burst & Avg. & P-P & Burst & Avg. & P-P & Burst & Avg. \\ \midrule
\multirow{3}{*}{\textbf{GPT-4}}      & 3-Turn   &   \textbf{+0.6}  &   -0.5    & \textbf{56.2} & 49.1  & \textbf{51.9} & 50  & 53.3  & 52   & \textbf{60}  & 53.3  & 56   \\
                                    & 10-Turn  &   +2.3  &   -0.8    & 40.5 & 37.9  & 38.9 & 40  & 46.7  & 44   & 50  & \underline{33.3}  & 40   \\
                                    & 110-Turn &   +3.5  &   +1.2   & 14.1 & 12.8  & 13.3 & 40  & 33.3  & 36   & \underline{10}  & 46.7  & \underline{32}   \\ \midrule
\multirow{3}{*}{\textbf{Claude-3-Sonnet}}      & 3-Turn   &  +2.1   &  \textbf{+0.4}   & 52.4 & \textbf{51.4}  & 51.8 & \textbf{60}  &  53.3 & 56  &  \textbf{60} &  46.7 &  52  \\
                                    & 10-Turn  &  +11.9  &  +1.5    & 19.0 & 40.0  & 32.1 &  40  &  53.3 & 48  &  40 &  53.3 &  48   \\
                                    & 110-Turn &  \underline{+16.9}  &  +4.5  & \underline{0} & 11.4  & 7.1 &  40  &  40 & 40  &  40  &  46.7 &   44   \\ \midrule
\multirow{3}{*}{\textbf{Qwen-110B}} & 3-Turn   &   +2.6  &   +3.8    & 48.2 & 45.3  & 44.6 & \textbf{60}  & \textbf{60}    & \textbf{60}   & \textbf{60}  & \textbf{73.3}  & \textbf{68}   \\
                                    & 10-Turn  &  +4.0   &   +3.9    & 20.5 & 24.2  & 22.7 & 50  & 40    & 44   & \textbf{60}  & 40    & 48 \\
                                    & 110-Turn &  +14.1   &   \underline{+7.6}    & \underline{0}    & \underline{1.8}   & \underline{1.1}  & \underline{30}  & \underline{33.3}  & \underline{32}   & 50  & 40    & 44   \\ \bottomrule
                                   
\end{tabular}}
\caption{Pass rate (\%) judged by human and LLMs. P-P stands for ping-pong dialogue, while Burst represents burst dialogue. \# Words denotes the average word count per message relative to the target individual. \textbf{Bold} underscores being viewed as most human-like, while \underline{underline} indicates the most deviation from real human responses.}
\label{tab:res}
\end{table*}

% Human: 3turn 12.9/8.25 10turn 13.0/8.4

\subsection{Test Setup}

We provide 100 turns of dialogue history in the context and evaluate the capabilities of GPT-4 \cite{DBLP:journals/corr/abs-2303-08774}, Claude-3-Sonnet \cite{TheC3} and Qwen-110B-Chat
%\footnote{https://huggingface.co/Qwen/Qwen1.5-110B-Chat} 
\cite{qwen} (referred to as Qwen-110B hereinafter) to respond like a human in real-life conversation across 25 topics including education, technology, art and more. The topic distribution is illustrated in Figure \ref{fig:topic}. These topics are carefully selected to cover a broad range of scenarios where conversational AI can be applied while avoiding appearing in the dialogue history. We split 10 topics for ping-pong dialogue and 15 for burst dialogue. Without loss of generality, we set $m$ and $m_T$ to 10.

\paragraph{Human Judges}
In total, 236 native speakers participated in our survey as volunteer human judges. These judges are recruited from diverse backgrounds to ensure a wide range of perspectives, tasked with evaluating pairs of conversations and distinguishing the human-human conversation from the human-machine one.

\paragraph{LLM Judges} 
Exploring whether LLM can distinguish between humans and itself can also provide insights into its understanding of social interactions among humans. We select GPT-4 and Qwen-110B to serve as LLM judges. To mitigate the effects of position bias \cite{Zheng2023JudgingLW}, we randomly assign the positions of the interaction pieces during the evaluation.

\subsection{Metric: X-Turn Pass Rate}

In the evolving landscape of human-AI interaction, the ability of an LLM to sustain human-like conversation over an arbitrary number of turns can be more revealing when assessing its human likeness. Thus, we propose the \textit{X-turn Pass Rate} metric to evaluate the possibility that the model can pass the Turing test within specified turns of dialogue. To compute the metric over $N$ pairs of $X$-turn dialogues, we ask $K$ judges (with $K=1$ for each LLM judge) to differentiate between each pair, comprising a human-human and a human-machine conversation. We record the number of judges who accurately identify the human-machine conversation in each pair as $C = \{C_1, C_2, \dots, C_N\}$, where $C_i$ represents the count for the $i$-th pair. The pass rate for the model in $X$ turns is then calculated by:
\begin{equation*}
\text{Pass Rate} = 1 - \frac{1}{N} \sum_{i=1}^N \left(\frac{C_i}{K}\right)
\end{equation*}
This metric evaluates the proportion of dialogues in which the judges fail to distinguish between human and machine. A higher X-turn pass rate signifies more human-like performance by the LLM, indicating greater success in the test.

\subsection{Results}

Table \ref{tab:res} shows the pass rates under different dialogue settings and judges. The 110-turn Turing test is conducted with a sequence of 100-turn pseudo-dialogue generation followed by a 10-turn interaction between human and machine. The results highlight a significant influence of dialogue length on model performance. According to the human judges, GPT-4 maintains a consistently higher average pass rate across all tested turns compared with other models, achieving the best average 3-turn pass rate of 51.9\%. However, as the dialogue progresses, the average word count per message has significantly increased and both models generally exhibited a significant decrease in their pass rates, showcasing the necessity of long Turing tests. 

\begin{figure*}[ht]
    \centering
    \includegraphics[width=0.95\linewidth]{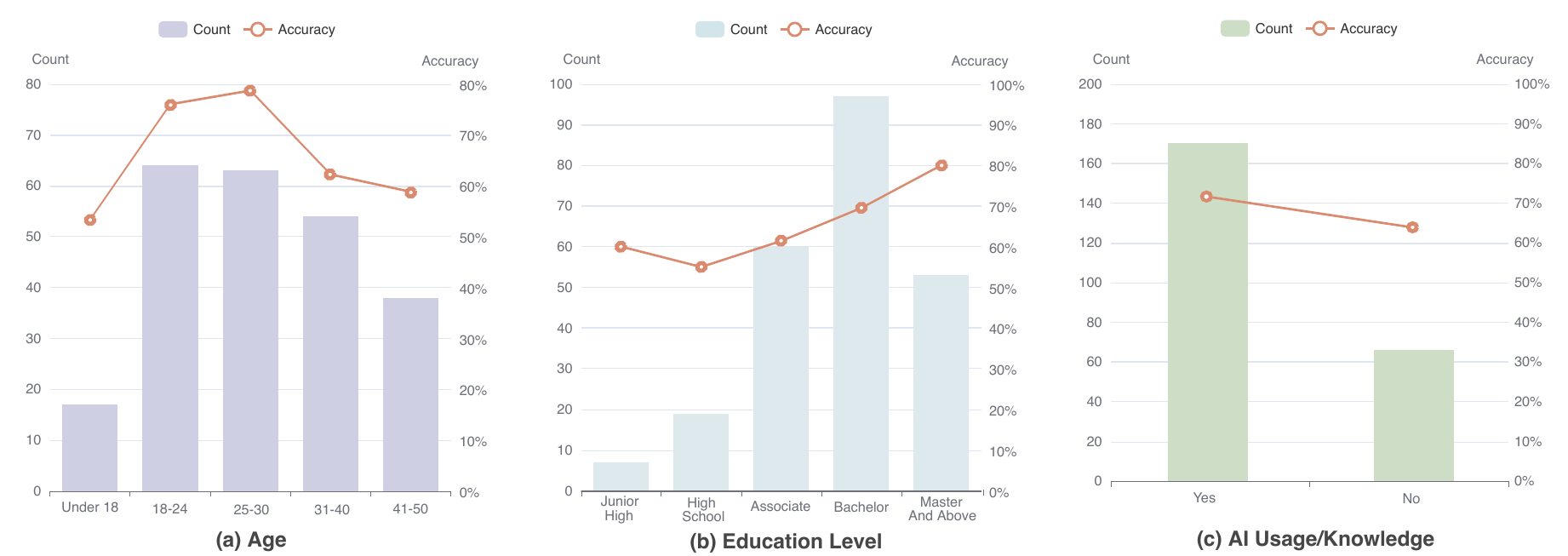}
    \caption{Participant distribution across different demographics, showcasing age, education level, and AI knowledge, highlighting their average test accuracy in the 10-turn Turing test.}
    \label{fig:discussion}
\end{figure*}

\begin{table}[t]
\centering
\small
\setlength{\tabcolsep}{1mm}{
\begin{tabular}{ccccc}
\toprule
\textbf{Model} & \textbf{\# Dia.}& \textbf{P-P} & \textbf{Burst} & \textbf{Avg.} \\ \midrule
\multirow{3}{*}{\textbf{GPT-4}}     & 10  & 35.3          & 31.2          & 33.4          \\
                                    & 50  & \textbf{41.2} & 35.2          & 37.4          \\
                                    & 100 & 40.5          & \textbf{37.9} & \textbf{38.9} \\ \midrule
\multirow{3}{*}{\textbf{Claude-3-Sonnet}}     & 10  &      14.3     &     37.1      & 28.6          \\
                                    & 50  & \textbf{19.0} &     37.1      &     30.4      \\
                                    & 100 & \textbf{19.0}  & \textbf{40.0} & \textbf{32.1} \\ \midrule
\multirow{3}{*}{\textbf{Qwen-110B}} & 10  & 17.0          & 18.2          & 17.7          \\
                                    & 50  & 20.3          & 20.2          & 20.2          \\
                                    & 100 & \textbf{20.5} & \textbf{24.2} & \textbf{22.7} \\ \bottomrule
\end{tabular}}
\caption{10-turn pass rate (\%) given different amount of dialogue history judged by humans.}
\label{tab:history}
\end{table}

\paragraph{Gap Between Human and LLM Judgment}
As the number of testing turns increases, all judges become increasingly adept at distinguishing between LLM and human interlocutors, resulting in a decreasing pass rate with an increased number of turns. However, the disparity between human and AI evaluation is evident. Human judges generally perceive GPT-4 and Claude-3-Sonnet as more human-like in conversations, whereas both LLMs tend to prefer Qwen-110B. Moreover, when presented with identical dialogues, human judges are generally more accurate in differentiating between LLM and human, contributing to a lower pass rate. This gap highlights the insufficiency of models in understanding and simulating real-world communications. Consequently, analysis in subsequent sections will primarily rely on assessments from human judges.

\paragraph{Comparison Between Dialogue Settings}
Comparing the extended burst dialogue setting with the classical ping-pong setting reveals notable differences. Although characterized by more complex information exchanges, all models better mimic human messaging habits in the number of words per message under the burst setting. However, performance variations exist between the two patterns, with Claude-3-Sonnet and Qwen-110B excelling in burst dialogues, while GPT-4 performs better in ping-pong dialogues. Therefore, as a natural form of textual communication in human life, challenges presented by burst dialogue are also worthy of attention.

\section{Discussion}

In this section, we investigate other significant factors that have an impact on the \textbf{\textsc{X-Turing}} and ultimate evaluation results.

\subsection{Dialogue History Quantity}

\begin{figure*}[ht]
    \centering
    \includegraphics[width=\linewidth]{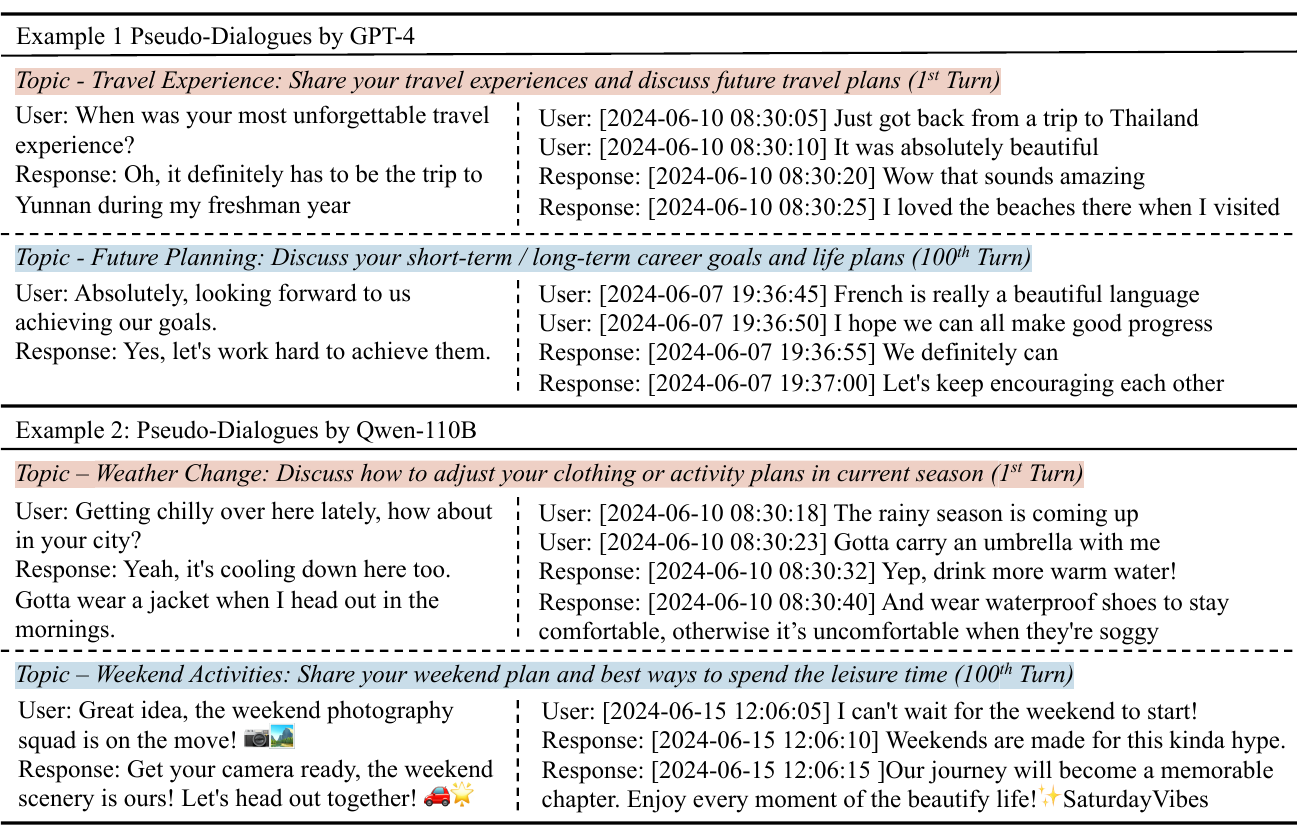}
    \caption{Case study of generated pseudo-dialogues at different turns of iteration. The deviation in dialogue style between the first and the 100th turn of Qwen-110B demonstrates that pseudo-dialogue generation can reflect the long-term dialogue agent's ability to consistently simulate the role in the instructions.}
    \label{fig:pseudo-dialogue}
\end{figure*}

A pivotal consideration of having LLM play a specific role in giving utterances is determining the necessary amount of utterances. To explore this, we analyze how the quantity of dialogue history provided for the model affects the pass rates of the Turing test, as presented in Table \ref{tab:history}. Generally, providing more dialogue history enables models to better simulate human conversational styles, thereby increasing their pass rate. This effect is particularly pronounced in burst dialogue, where additional dialogue examples are markedly effective. However, the benefit of increasing the given dialogue diminishes once a certain threshold is surpassed. Moreover, the consistency of the dialogue style plays a crucial role. Redundant or low-quality data can lead to anomalies, such as GPT-4 performing better given 50 turns of ping-pong dialogue data compared to 100 turns. This highlights the importance of both the quantity and quality of dialogue history in effectively helping LLMs through the Turing Test.

\subsection{Participant Distribution}

Focusing on the 10-turn Turing test, we analyze how participant distribution impacts the evaluation. Figure \ref{fig:discussion} provides a comprehensive overview of participant demographics and their corresponding accuracy in the Turing Test: (1) \textbf{Age Impact}: The age group distribution reveals varied levels of accuracy in distinguishing AI from human participants, with the age groups 18-24 and 25-30 achieving the highest accuracy rates. It highlights the influence of age on technological adaptability and cognitive acuity. (2) \textbf{Educational Influence}: There is a noticeable trend where participants with higher educational levels exhibit higher accuracy rates. This suggests that educational background may play a crucial role in understanding and interacting with AI, possibly due to better critical thinking skills and familiarity with technology. (3) \textbf{AI Usage and Knowledge}: Both groups of participants can distinguish between human and AI responses with moderate accuracy, even those who claim to have no experience with AI. As anticipated, participants who report previous AI usage or knowledge demonstrate higher accuracy.

\subsection{Topic Analysis}

\begin{figure}[t]
    \centering
    \includegraphics[width=0.95\linewidth]{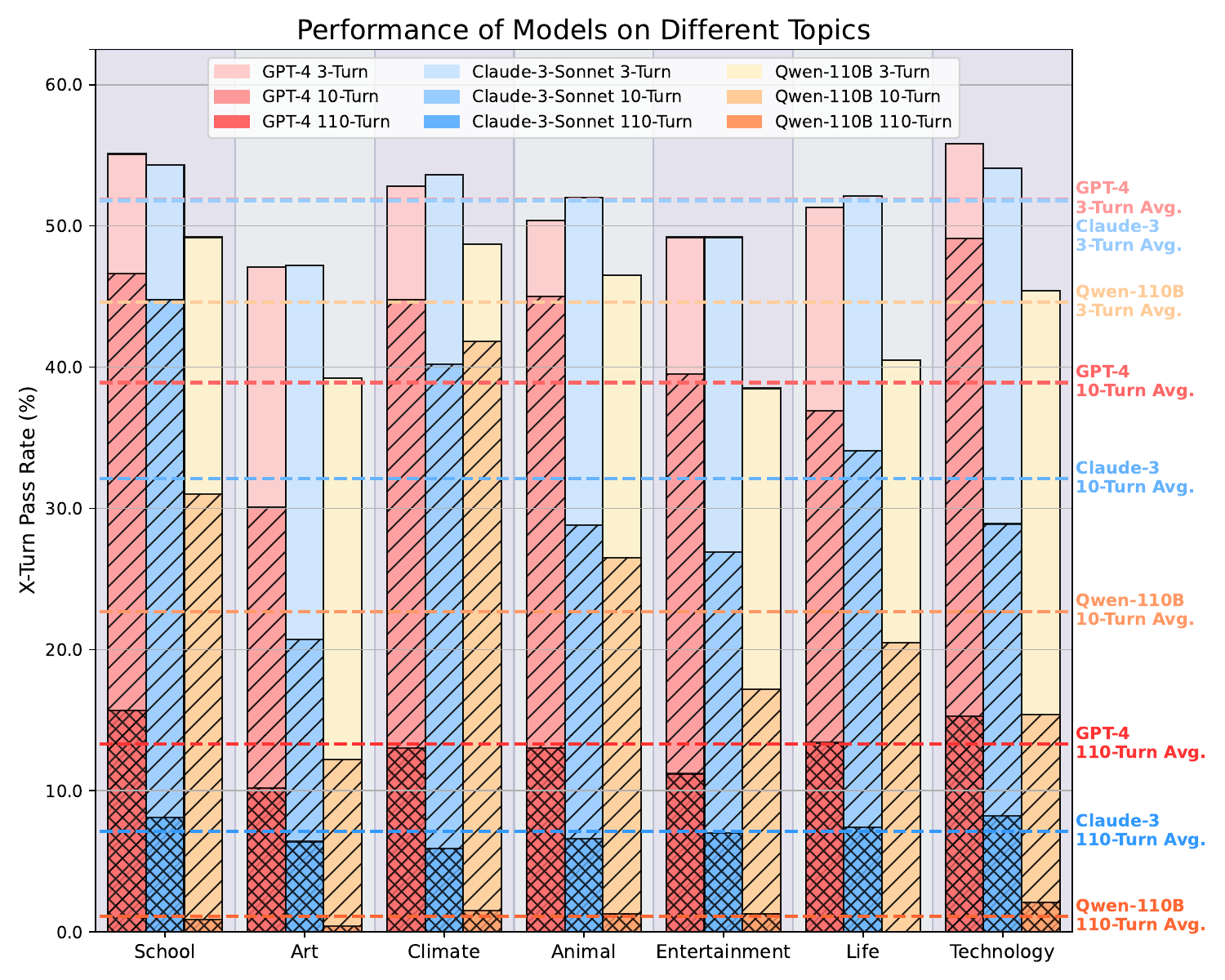}
    \caption{The 3/10/110-Turn Pass Rate of experimented models across the tested topics.}
    \label{fig:topic-impact}
\end{figure}

Considering the varying amounts and qualities of data in the pretraining corpora for different topics, the capacity of models to generate contextually appropriate responses could be different. It is necessary to investigate the impact of topics on the test while ensuring a comprehensive and balanced distribution. As shown in Figure \ref{fig:topic-impact}, there is an evident disparity in how different models perform on distinct topics at shorter dialogue duration (3-turn and 10-turn). For instance, GPT-4 excels in Technology, while Claude-3-Sonnet shows a higher aptitude in School-related topics and Qwen-110B has a relative strength in Climate discussions. However, a common challenge among all three models is observed in the Art category, suggesting that the creative and interpretive nature of artistic discussions presents a difficulty for LLMs in achieving human-level conversation. However, as dialogue lengthens to 100 turns, the overall average performance declines and the ability to sustain consistent interaction becomes a more dominant factor, overshadowing the advantages or disadvantages that particular topics might offer.

\section{Case Study}

\subsection{Pseudo-Dialogue Samples}

We examine the capabilities of LLMs in maintaining consistency and 
naturalness in generating pseudo-dialogues, using GPT-4 and Qwen-110B as examples. This assessment reflects the expected quality of subsequent human-machine interactions. As presented in Figure \ref{fig:pseudo-dialogue}, GPT-4 exhibits robustness in long dialogues, showing minimal degradation in the quality of language style imitation in its responses. Nonetheless, the model displays a reduced initiative in engaging with topics, with responses predominantly echoing agreement rather than fostering deep discussions. While for Qwen-110B, there is a noticeable shift in its conversational style by the 100th turn, marked by increasingly verbose replies and unconventional emoticon use, diverging from typical human conversational patterns. The comparative analysis emphasizes GPT-4's enhanced ability to preserve human-like qualities over prolonged interactions compared to Qwen-110B, despite still failing to deceive humans effectively. This finding aligns with the human evaluation results presented in Table \ref{tab:res}.

\begin{figure}[t]
    \centering
    \includegraphics[width=0.95\linewidth]{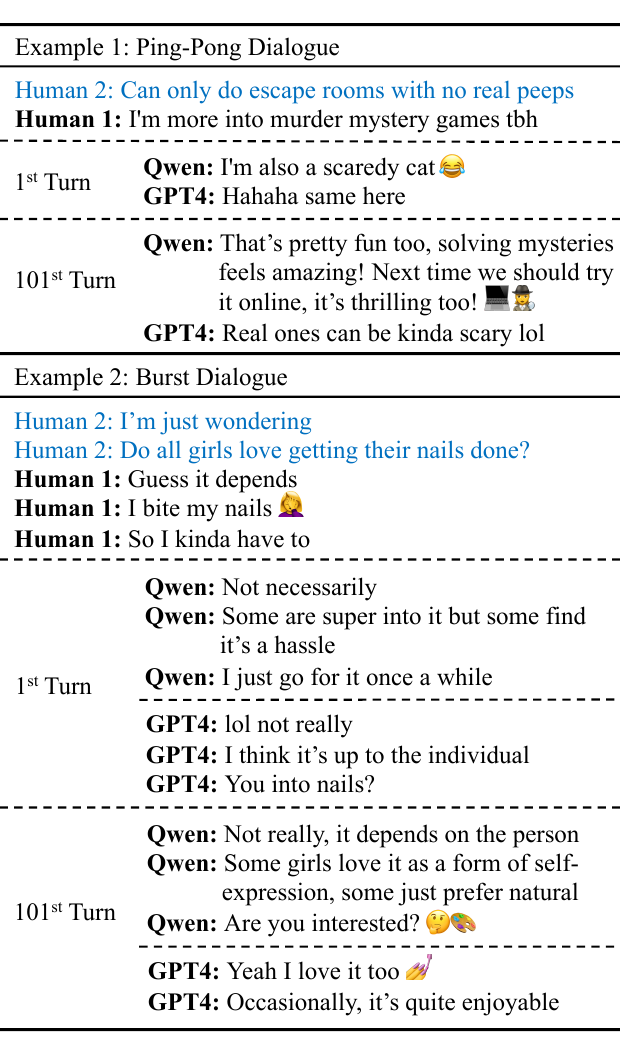}
    \caption{Case study of dialogues at first and one hundred first turn during \textbf{\textsc{X-Turing}}.}
    \label{fig:interaction}
\end{figure}

\subsection{Interaction Samples}

Figure \ref{fig:interaction} highlights the differences between AI and human responses at the first and 101st turn (after 100 pseudo-dialogue turns) in two distinct dialogue settings. In both dialogue types, the LLMs initially mirror human conversational tactics effectively, utilizing informal slang and abbreviations, and maintaining brevity. However, as the conversations progress to the 101st turn, notable discrepancies begin to appear. Notably, in the Ping-Pong dialogue format, Qwen begins to produce overly lengthy responses, a deviation from the concise exchanges typically observed in natural human interactions. This extended message length makes the conversation feel less natural and more mechanical, as reported by most of the judges. 
Moreover, while the use of emoticons by LLMs aims to enhance emotional expressiveness, it often misses the mark in its execution. 
Rather than expressing emotions with emoticons as how they are typically employed in human text conversations, responses by LLMs like Qwen-110B often feature an overuse of emojis that summarize the textual content. These issues underscore the ongoing challenges LLMs face in replicating the nuanced flow and emotional subtlety of human conversations.

\section{Conclusion}

In summary, our study introduces the \textbf{\textsc{X-Turing}} framework that effectively and efficiently evaluates the human likeness of LLMs through pseudo-dialogue generation with extended dialogue settings. With the definition of the \textit{X-Turn Pass-Rate} metric, our experiments reveal that GPT-4 outperforms other tested models across most configurations and evaluation criteria, achieving an average 51.9\% and 38.9\% pass rate in 3-turn and 10-turn dialogue respectively. However, as dialogue length increases through iteratively generating dialogues that approximate human-machine interactions, the pass rates for all models significantly drop. Moreover, burst dialogue presents a greater challenge in maintaining consistency and human-like responses. Our analysis highlights the critical influence of dialogue duration and complexity on LLMs' performance in Turing tests and suggests directions for refining evaluation methods to enhance efficiency and reliability.

\section*{Limitations}

While our study presents an innovative approach to efficiently evaluating LLMs using the extended Turing test framework, the following limitations must be acknowledged: (1) Responses from LLMs are inherently influenced by user inputs and previous LLM-generated dialogue quality. Hence, using static evaluation techniques like questionnaires instead of real-time communication cannot eliminate the impact of the preferences of users participating in the conversation, which forms a trade-off between efficiency and accuracy. (2) Although we considered various participant factors such as age, educational background, and familiarity with AI, the subjective nature of human judgment in evaluating AI responses can lead to inconsistencies, which is an intrinsic characteristic of the Turing test.

\section*{Ethics Statement}

Our research involves collecting dialogue data and questionnaires from real human participants, and as such, we adhere to stringent ethical guidelines to ensure the privacy, consent, and well-being of all participants involved. Key ethical considerations include: (1) Informed Consent: All participants are provided with detailed information about the study's purpose, procedures, and their rights. They can withdraw from the study at any time without any consequences. (2) Data Anonymization: To protect the privacy of participants, all collected dialogue data and questionnaires are anonymized. Personal identifiers are removed to ensure that individual participants cannot be traced from the data. (3) Data Security: Collected data and information are stored securely, and access is restricted to authorized personnel only. We employ robust data protection measures to prevent unauthorized access, disclosure, or misuse of the data.

\bibliography{anthology,custom}

\begin{thebibliography}{46}
\expandafter\ifx\csname natexlab\endcsname\relax\def\natexlab#1{#1}\fi

\bibitem[{Abbasiantaeb et~al.(2023)Abbasiantaeb, Yuan, Kanoulas, and Aliannejadi}]{Abbasiantaeb2023LetTL}
Zahra Abbasiantaeb, Yifei Yuan, E.~Kanoulas, and Mohammad Aliannejadi. 2023.
\newblock \href {https://api.semanticscholar.org/CorpusID:265659354} {Let the llms talk: Simulating human-to-human conversational qa via zero-shot llm-to-llm interactions}.
\newblock \emph{Proceedings of the 17th ACM International Conference on Web Search and Data Mining}.

\bibitem[{Abdullin et~al.(2024)Abdullin, Aliod, Ofoghi, Yearwood, and Li}]{Abdullin2024SyntheticDD}
Yelaman Abdullin, Diego~Moll{\'a} Aliod, Bahadorreza Ofoghi, John Yearwood, and Qingyang Li. 2024.
\newblock \href {https://api.semanticscholar.org/CorpusID:267335111} {Synthetic dialogue dataset generation using llm agents}.
\newblock \emph{ArXiv}, abs/2401.17461.

\bibitem[{Anthropic(2024)}]{TheC3}
Anthropic. 2024.
\newblock \href {https://www-cdn.anthropic.com/de8ba9b01c9ab7cbabf5c33b80b7bbc618857627/Model_Card_Claude_3.pdf} {The claude 3 model family: Opus, sonnet, haiku}.

\bibitem[{Bai et~al.(2023)Bai, Bai, Chu, Cui, Dang, Deng, Fan, Ge, Han, Huang, Hui, Ji, Li, Lin, Lin, Liu, Liu, Lu, Lu, Ma, Men, Ren, Ren, Tan, Tan, Tu, Wang, Wang, Wang, Wu, Xu, Xu, Yang, Yang, Yang, Yang, Yao, Yu, Yuan, Yuan, Zhang, Zhang, Zhang, Zhang, Zhou, Zhou, Zhou, and Zhu}]{qwen}
Jinze Bai, Shuai Bai, Yunfei Chu, Zeyu Cui, Kai Dang, Xiaodong Deng, Yang Fan, Wenbin Ge, Yu~Han, Fei Huang, Binyuan Hui, Luo Ji, Mei Li, Junyang Lin, Runji Lin, Dayiheng Liu, Gao Liu, Chengqiang Lu, Keming Lu, Jianxin Ma, Rui Men, Xingzhang Ren, Xuancheng Ren, Chuanqi Tan, Sinan Tan, Jianhong Tu, Peng Wang, Shijie Wang, Wei Wang, Shengguang Wu, Benfeng Xu, Jin Xu, An~Yang, Hao Yang, Jian Yang, Shusheng Yang, Yang Yao, Bowen Yu, Hongyi Yuan, Zheng Yuan, Jianwei Zhang, Xingxuan Zhang, Yichang Zhang, Zhenru Zhang, Chang Zhou, Jingren Zhou, Xiaohuan Zhou, and Tianhang Zhu. 2023.
\newblock \href {https://arxiv.org/pdf/2309.16609} {Qwen technical report}.
\newblock \emph{arXiv preprint arXiv:2309.16609}.

\bibitem[{Boyd et~al.(2020)Boyd, Puri, Shoeybi, Patwary, and Catanzaro}]{Boyd2020LargeSM}
Alex Boyd, Raul Puri, Mohammad Shoeybi, Mostofa Patwary, and Bryan Catanzaro. 2020.
\newblock \href {https://api.semanticscholar.org/CorpusID:218613714} {Large scale multi-actor generative dialog modeling}.
\newblock \emph{ArXiv}, abs/2005.06114.

\bibitem[{Bubeck et~al.(2023)Bubeck, Chandrasekaran, Eldan, Gehrke, Horvitz, Kamar, Lee, Lee, Li, Lundberg, Nori, Palangi, Ribeiro, and Zhang}]{Bubeck2023SparksOA}
S{\'e}bastien Bubeck, Varun Chandrasekaran, Ronen Eldan, John~A. Gehrke, Eric Horvitz, Ece Kamar, Peter Lee, Yin~Tat Lee, Yuan-Fang Li, Scott~M. Lundberg, Harsha Nori, Hamid Palangi, Marco~Tulio Ribeiro, and Yi~Zhang. 2023.
\newblock \href {https://api.semanticscholar.org/CorpusID:257663729} {Sparks of artificial general intelligence: Early experiments with gpt-4}.
\newblock \emph{ArXiv}, abs/2303.12712.

\bibitem[{Chen et~al.(2023{\natexlab{a}})Chen, Papangelis, Tao, Kim, Rosenbaum, Liu, Yu, and Hakkani-T{\"u}r}]{Chen2023PLACESPL}
Maximillian Chen, Alexandros Papangelis, Chenyang Tao, Seokhwan Kim, Andrew Rosenbaum, Yang Liu, Zhou Yu, and Dilek~Z. Hakkani-T{\"u}r. 2023{\natexlab{a}}.
\newblock \href {https://api.semanticscholar.org/CorpusID:256627323} {Places: Prompting language models for social conversation synthesis}.
\newblock \emph{ArXiv}, abs/2302.03269.

\bibitem[{Chen et~al.(2023{\natexlab{b}})Chen, Yu, Shi, Awasthi, and Yu}]{Chen2023ControllableMD}
Maximillian Chen, Xiao Yu, Weiyan Shi, Urvi Awasthi, and Zhou Yu. 2023{\natexlab{b}}.
\newblock \href {https://api.semanticscholar.org/CorpusID:258557267} {Controllable mixed-initiative dialogue generation through prompting}.
\newblock \emph{ArXiv}, abs/2305.04147.

\bibitem[{Chen et~al.(2023{\natexlab{c}})Chen, Wang, Jiang, Cai, Li, Chen, Wang, and Li}]{chen-etal-2023-large}
Nuo Chen, Yan Wang, Haiyun Jiang, Deng Cai, Yuhan Li, Ziyang Chen, Longyue Wang, and Jia Li. 2023{\natexlab{c}}.
\newblock \href {https://doi.org/10.18653/v1/2023.findings-emnlp.570} {Large language models meet harry potter: A dataset for aligning dialogue agents with characters}.
\newblock In \emph{Findings of the Association for Computational Linguistics: EMNLP 2023}, pages 8506--8520, Singapore. Association for Computational Linguistics.

\bibitem[{Chen et~al.(2024)Chen, Nishida, Nakayama, and Matsumoto}]{Chen2024RecentTI}
Yi-Pei Chen, Noriki Nishida, Hideki Nakayama, and Yuji Matsumoto. 2024.
\newblock \href {https://api.semanticscholar.org/CorpusID:269804123} {Recent trends in personalized dialogue generation: A review of datasets, methodologies, and evaluations}.
\newblock In \emph{International Conference on Language Resources and Evaluation}.

\bibitem[{Freitas et~al.(2020)Freitas, Luong, So, Hall, Fiedel, Thoppilan, Yang, Kulshreshtha, Nemade, Lu, and Le}]{DeFreitas2020TowardsAH}
Daniel~De Freitas, Minh-Thang Luong, David~R. So, Jamie Hall, Noah Fiedel, Romal Thoppilan, Zi~Yang, Apoorv Kulshreshtha, Gaurav Nemade, Yifeng Lu, and Quoc~V. Le. 2020.
\newblock \href {https://api.semanticscholar.org/CorpusID:210920238} {Towards a human-like open-domain chatbot}.
\newblock \emph{ArXiv}, abs/2001.09977.

\bibitem[{Gao and Emami(2023)}]{Gao2023TheTQ}
Qi~Chen Gao and Ali Emami. 2023.
\newblock \href {https://api.semanticscholar.org/CorpusID:259370766} {The turing quest: Can transformers make good npcs?}
\newblock In \emph{Annual Meeting of the Association for Computational Linguistics}.

\bibitem[{Han et~al.(2022)Han, Kim, Yoo, Seo, Kim, Erdenee, and Chang}]{han-etal-2022-meet}
Seungju Han, Beomsu Kim, Jin~Yong Yoo, Seokjun Seo, Sangbum Kim, Enkhbayar Erdenee, and Buru Chang. 2022.
\newblock \href {https://doi.org/10.18653/v1/2022.naacl-main.377} {Meet your favorite character: Open-domain chatbot mimicking fictional characters with only a few utterances}.
\newblock In \emph{Proceedings of the 2022 Conference of the North American Chapter of the Association for Computational Linguistics: Human Language Technologies}, pages 5114--5132, Seattle, United States. Association for Computational Linguistics.

\bibitem[{Helm et~al.(2023)Helm, Priebe, and Yang}]{Helm2023AST}
Hayden~S. Helm, Carey~E. Priebe, and Weiwei Yang. 2023.
\newblock \href {https://api.semanticscholar.org/CorpusID:262043684} {A statistical turing test for generative models}.
\newblock \emph{ArXiv}, abs/2309.08913.

\bibitem[{Imanpour et~al.(2024)Imanpour, Bajpai, Ghosh, Sankepally, Borah, Abdullah, Kosaraju, Dixit, Aziz, Biswas, Jain, Chadha, Sheth, and Das}]{imanpour2024visualcounterturingtest}
Nasrin Imanpour, Shashwat Bajpai, Subhankar Ghosh, Sainath~Reddy Sankepally, Abhilekh Borah, Hasnat~Md Abdullah, Nishoak Kosaraju, Shreyas Dixit, Ashhar Aziz, Shwetangshu Biswas, Vinija Jain, Aman Chadha, Amit Sheth, and Amitava Das. 2024.
\newblock \href {http://arxiv.org/abs/2411.16754} {Visual counter turing test (vct\^{}2): Discovering the challenges for ai-generated image detection and introducing visual ai index (v\_ai)}.

\bibitem[{Jandaghi et~al.(2023)Jandaghi, Sheng, Bai, Pujara, and Sidahmed}]{Jandaghi2023FaithfulPC}
Pegah Jandaghi, XiangHai Sheng, Xinyi Bai, Jay Pujara, and Hakim Sidahmed. 2023.
\newblock \href {https://api.semanticscholar.org/CorpusID:266335707} {Faithful persona-based conversational dataset generation with large language models}.
\newblock \emph{ArXiv}, abs/2312.10007.

\bibitem[{Jannai et~al.(2023)Jannai, Meron, Lenz, Levine, and Shoham}]{Jannai2023HumanON}
Daniel Jannai, Amos Meron, Barak Lenz, Yoav Levine, and Yoav Shoham. 2023.
\newblock \href {https://api.semanticscholar.org/CorpusID:258987666} {Human or not? a gamified approach to the turing test}.
\newblock \emph{ArXiv}, abs/2305.20010.

\bibitem[{Jiang et~al.(2023)Jiang, Sablayrolles, Mensch, Bamford, Chaplot, de~Las~Casas, Bressand, Lengyel, Lample, Saulnier, Lavaud, Lachaux, Stock, Scao, Lavril, Wang, Lacroix, and Sayed}]{DBLP:journals/corr/abs-2310-06825}
Albert~Q. Jiang, Alexandre Sablayrolles, Arthur Mensch, Chris Bamford, Devendra~Singh Chaplot, Diego de~Las~Casas, Florian Bressand, Gianna Lengyel, Guillaume Lample, Lucile Saulnier, L{\'{e}}lio~Renard Lavaud, Marie{-}Anne Lachaux, Pierre Stock, Teven~Le Scao, Thibaut Lavril, Thomas Wang, Timoth{\'{e}}e Lacroix, and William~El Sayed. 2023.
\newblock \href {https://doi.org/10.48550/ARXIV.2310.06825} {Mistral 7b}.
\newblock \emph{CoRR}, abs/2310.06825.

\bibitem[{Jones and Bergen(2024)}]{Jones2024PeopleCD}
Cameron~R. Jones and Benjamin~K. Bergen. 2024.
\newblock \href {https://api.semanticscholar.org/CorpusID:269761698} {People cannot distinguish gpt-4 from a human in a turing test}.

\bibitem[{Kim et~al.(2022{\natexlab{a}})Kim, Hessel, Jiang, West, Lu, Yu, Zhou, Bras, Alikhani, Kim, Sap, and Choi}]{Kim2022SODAMD}
Hyunwoo Kim, Jack Hessel, Liwei Jiang, Peter West, Ximing Lu, Youngjae Yu, Pei Zhou, Ronan~Le Bras, Malihe Alikhani, Gunhee Kim, Maarten Sap, and Yejin Choi. 2022{\natexlab{a}}.
\newblock \href {https://api.semanticscholar.org/CorpusID:254877312} {Soda: Million-scale dialogue distillation with social commonsense contextualization}.
\newblock In \emph{Conference on Empirical Methods in Natural Language Processing}.

\bibitem[{Kim et~al.(2022{\natexlab{b}})Kim, Kim, Song, won Hwang, and Yeo}]{Kim2022BotsTalkMF}
Minju Kim, Chaehyeong Kim, Yongho Song, Seung won Hwang, and Jinyoung Yeo. 2022{\natexlab{b}}.
\newblock \href {https://api.semanticscholar.org/CorpusID:249364288} {Botstalk: Machine-sourced framework for automatic curation of large-scale multi-skill dialogue datasets}.
\newblock In \emph{Conference on Empirical Methods in Natural Language Processing}.

\bibitem[{Lee et~al.(2023)Lee, Heo, Choi, Choi, and Zhang}]{VideoTuringTest2023}
Minsu Lee, Yu‐Jung Heo, Seongho Choi, Woo Choi, and Byoung-Tak Zhang. 2023.
\newblock \href {https://doi.org/10.1002/aaai.12128} {Video turing test: A first step towards human‐level ai}.
\newblock \emph{AI Magazine}, 44.

\bibitem[{Lee et~al.(2022)Lee, Lim, Choi, Lm, and Choi}]{Lee2022PERSONACHATGENGP}
Young-Jun Lee, Chae-Gyun Lim, Yunsu Choi, Ji-Hui Lm, and Ho-Jin Choi. 2022.
\newblock \href {https://api.semanticscholar.org/CorpusID:252819096} {Personachatgen: Generating personalized dialogues using gpt-3}.
\newblock In \emph{CCGPK}.

\bibitem[{Li et~al.(2023)Li, Leng, Yan, Shen, Wang, MI, Fei, Feng, Yan, Wang, Zhan, Jia, Wu, and Sun}]{li2023chatharuhi}
Cheng Li, Ziang Leng, Chenxi Yan, Junyi Shen, Hao Wang, Weishi MI, Yaying Fei, Xiaoyang Feng, Song Yan, HaoSheng Wang, Linkang Zhan, Yaokai Jia, Pingyu Wu, and Haozhen Sun. 2023.
\newblock \href {http://arxiv.org/abs/2308.09597} {Chatharuhi: Reviving anime character in reality via large language model}.

\bibitem[{Lowe et~al.(2017)Lowe, Noseworthy, Serban, Angelard-Gontier, Bengio, and Pineau}]{Lowe2017TowardsAA}
Ryan Lowe, Michael Noseworthy, Iulian Serban, Nicolas Angelard-Gontier, Yoshua Bengio, and Joelle Pineau. 2017.
\newblock \href {https://api.semanticscholar.org/CorpusID:1880070} {Towards an automatic turing test: Learning to evaluate dialogue responses}.
\newblock In \emph{Annual Meeting of the Association for Computational Linguistics}.

\bibitem[{Lu et~al.(2024)Lu, Yu, Zhou, and Zhou}]{lu2024large}
Keming Lu, Bowen Yu, Chang Zhou, and Jingren Zhou. 2024.
\newblock \href {http://arxiv.org/abs/2401.12474} {Large language models are superpositions of all characters: Attaining arbitrary role-play via self-alignment}.

\bibitem[{Murdock and Bennet(1962)}]{Murdock1962TheSP}
Murdock and B~Bennet. 1962.
\newblock \href {https://api.semanticscholar.org/CorpusID:6752448} {The serial position effect of free recall}.
\newblock \emph{Journal of Experimental Psychology}, 64:482--488.

\bibitem[{Ng et~al.(2024)Ng, Tse, tse Huang, Li, Wang, and Lyu}]{Ng2024HowWC}
Man~Tik Ng, Hui~Tung Tse, Jen tse Huang, Jingjing Li, Wenxuan Wang, and Michael~R. Lyu. 2024.
\newblock \href {https://api.semanticscholar.org/CorpusID:269293628} {How well can llms echo us? evaluating ai chatbots' role-play ability with echo}.
\newblock \emph{ArXiv}, abs/2404.13957.

\bibitem[{OpenAI(2023)}]{DBLP:journals/corr/abs-2303-08774}
OpenAI. 2023.
\newblock \href {https://doi.org/10.48550/arXiv.2303.08774} {{GPT-4} technical report}.
\newblock \emph{CoRR}, abs/2303.08774.

\bibitem[{Park et~al.(2023)Park, O'Brien, Cai, Morris, Liang, and Bernstein}]{Park2023GenerativeAI}
Joon~Sung Park, Joseph~C. O'Brien, Carrie~J. Cai, Meredith~Ringel Morris, Percy Liang, and Michael~S. Bernstein. 2023.
\newblock \href {https://api.semanticscholar.org/CorpusID:258040990} {Generative agents: Interactive simulacra of human behavior}.
\newblock \emph{Proceedings of the 36th Annual ACM Symposium on User Interface Software and Technology}.

\bibitem[{Shanahan et~al.(2023)Shanahan, McDonell, and Reynolds}]{Shanahan2023RolePW}
Murray Shanahan, Kyle McDonell, and Laria Reynolds. 2023.
\newblock \href {https://api.semanticscholar.org/CorpusID:258947657} {Role play with large language models}.
\newblock \emph{Nature}, 623:493--498.

\bibitem[{Shao et~al.(2023)Shao, Li, Dai, and Qiu}]{shao-etal-2023-character}
Yunfan Shao, Linyang Li, Junqi Dai, and Xipeng Qiu. 2023.
\newblock \href {https://aclanthology.org/2023.emnlp-main.814} {Character-{LLM}: A trainable agent for role-playing}.
\newblock In \emph{Proceedings of the 2023 Conference on Empirical Methods in Natural Language Processing}, pages 13153--13187, Singapore. Association for Computational Linguistics.

\bibitem[{Soudani et~al.(2024)Soudani, Petcu, Kanoulas, and Hasibi}]{Soudani2024ASO}
Heydar Soudani, Roxana Petcu, E.~Kanoulas, and Faegheh Hasibi. 2024.
\newblock \href {https://api.semanticscholar.org/CorpusID:269982301} {A survey on recent advances in conversational data generation}.

\bibitem[{Tao et~al.(2023)Tao, Liang, Shi, Yu, and Xie}]{tao2023rolecraftglm}
Meiling Tao, Xuechen Liang, Tianyu Shi, Lei Yu, and Yiting Xie. 2023.
\newblock \href {http://arxiv.org/abs/2401.09432} {Rolecraft-glm: Advancing personalized role-playing in large language models}.

\bibitem[{Touvron et~al.(2023)Touvron, Martin, Stone, Albert, Almahairi, Babaei, Bashlykov, Batra, Bhargava, Bhosale, Bikel, Blecher, Canton{-}Ferrer, Chen, Cucurull, Esiobu, Fernandes, Fu, Fu, Fuller, Gao, Goswami, Goyal, Hartshorn, Hosseini, Hou, Inan, Kardas, Kerkez, Khabsa, Kloumann, Korenev, Koura, Lachaux, Lavril, Lee, Liskovich, Lu, Mao, Martinet, Mihaylov, Mishra, Molybog, Nie, Poulton, Reizenstein, Rungta, Saladi, Schelten, Silva, Smith, Subramanian, Tan, Tang, Taylor, Williams, Kuan, Xu, Yan, Zarov, Zhang, Fan, Kambadur, Narang, Rodriguez, Stojnic, Edunov, and Scialom}]{DBLP:journals/corr/abs-2307-09288}
Hugo Touvron, Louis Martin, Kevin Stone, Peter Albert, Amjad Almahairi, Yasmine Babaei, Nikolay Bashlykov, Soumya Batra, Prajjwal Bhargava, Shruti Bhosale, Dan Bikel, Lukas Blecher, Cristian Canton{-}Ferrer, Moya Chen, Guillem Cucurull, David Esiobu, Jude Fernandes, Jeremy Fu, Wenyin Fu, Brian Fuller, Cynthia Gao, Vedanuj Goswami, Naman Goyal, Anthony Hartshorn, Saghar Hosseini, Rui Hou, Hakan Inan, Marcin Kardas, Viktor Kerkez, Madian Khabsa, Isabel Kloumann, Artem Korenev, Punit~Singh Koura, Marie{-}Anne Lachaux, Thibaut Lavril, Jenya Lee, Diana Liskovich, Yinghai Lu, Yuning Mao, Xavier Martinet, Todor Mihaylov, Pushkar Mishra, Igor Molybog, Yixin Nie, Andrew Poulton, Jeremy Reizenstein, Rashi Rungta, Kalyan Saladi, Alan Schelten, Ruan Silva, Eric~Michael Smith, Ranjan Subramanian, Xiaoqing~Ellen Tan, Binh Tang, Ross Taylor, Adina Williams, Jian~Xiang Kuan, Puxin Xu, Zheng Yan, Iliyan Zarov, Yuchen Zhang, Angela Fan, Melanie Kambadur, Sharan Narang, Aur{\'{e}}lien Rodriguez, Robert Stojnic, Sergey Edunov,
  and Thomas Scialom. 2023.
\newblock \href {https://doi.org/10.48550/arXiv.2307.09288} {Llama 2: Open foundation and fine-tuned chat models}.
\newblock \emph{CoRR}, abs/2307.09288.

\bibitem[{Turing(1950)}]{Turing1950ComputingMA}
Alan~M. Turing. 1950.
\newblock \href {https://api.semanticscholar.org/CorpusID:14636783} {Computing machinery and intelligence}.
\newblock \emph{Mind}, LIX:433--460.

\bibitem[{Uchendu et~al.(2021)Uchendu, Ma, Le, Zhang, and Lee}]{Uchendu2021TURINGBENCHAB}
Adaku Uchendu, Zeyu Ma, Thai Le, Rui Zhang, and Dongwon Lee. 2021.
\newblock \href {https://api.semanticscholar.org/CorpusID:237589233} {Turingbench: A benchmark environment for turing test in the age of neural text generation}.
\newblock In \emph{Conference on Empirical Methods in Natural Language Processing}.

\bibitem[{Wang et~al.(2023{\natexlab{a}})Wang, Tu, Fei, Leng, and Li}]{wang2023does}
Xintao Wang, Quan Tu, Yaying Fei, Ziang Leng, and Cheng Li. 2023{\natexlab{a}}.
\newblock \href {http://arxiv.org/abs/2310.17976} {Does role-playing chatbots capture the character personalities? assessing personality traits for role-playing chatbots}.

\bibitem[{Wang et~al.(2023{\natexlab{b}})Wang, Peng, Que, Liu, Zhou, Wu, Guo, Gan, Ni, Zhang, Zhang, Ouyang, Xu, Chen, Fu, and Peng}]{Wang2023RoleLLMBE}
Zekun Wang, Zhongyuan Peng, Haoran Que, Jiaheng Liu, Wangchunshu Zhou, Yuhan Wu, Hongcheng Guo, Ruitong Gan, Zehao Ni, Man Zhang, Zhaoxiang Zhang, Wanli Ouyang, Ke~Xu, Wenhu Chen, Jie Fu, and Junran Peng. 2023{\natexlab{b}}.
\newblock \href {https://api.semanticscholar.org/CorpusID:263334495} {Rolellm: Benchmarking, eliciting, and enhancing role-playing abilities of large language models}.
\newblock \emph{ArXiv}, abs/2310.00746.

\bibitem[{Wu et~al.(2024{\natexlab{a}})Wu, Xu, Xu, Hong, Wu, Zhao, Zhang, and He}]{wu2024scalingvirtualworlddeltaengine}
Hongqiu Wu, Zekai Xu, Tianyang Xu, Jiale Hong, Weiqi Wu, Hai Zhao, Min Zhang, and Zhezhi He. 2024{\natexlab{a}}.
\newblock \href {http://arxiv.org/abs/2408.05842} {Scaling virtual world with delta-engine}.

\bibitem[{Wu et~al.(2024{\natexlab{b}})Wu, Wu, Jiang, Liu, Zhao, and Zhang}]{wu-etal-2024-role}
Weiqi Wu, Hongqiu Wu, Lai Jiang, Xingyuan Liu, Hai Zhao, and Min Zhang. 2024{\natexlab{b}}.
\newblock \href {https://aclanthology.org/2024.findings-acl.196} {From role-play to drama-interaction: An {LLM} solution}.
\newblock In \emph{Findings of the Association for Computational Linguistics ACL 2024}, pages 3271--3290, Bangkok, Thailand and virtual meeting. Association for Computational Linguistics.

\bibitem[{Yang et~al.(2023)Yang, Xiao, Wang, Zhang, Bian, Yin, Lv, Pan, Wang, Yan, Yang, Deng, Wang, Liu, Ai, Dong, Zhao, Xu, Sun, Zhang, Liu, Ji, Xie, Dai, Fang, Su, Song, Liu, Ru, Ma, Wang, Liu, Lin, Nie, Guo, Sun, Zhang, Li, Li, Cheng, Chen, Zeng, Wang, Chen, Men, Yu, Pan, Shen, Wang, Li, Jiang, Gao, Zhang, Zhou, and Wu}]{DBLP:journals/corr/abs-2309-10305}
Aiyuan Yang, Bin Xiao, Bingning Wang, Borong Zhang, Ce~Bian, Chao Yin, Chenxu Lv, Da~Pan, Dian Wang, Dong Yan, Fan Yang, Fei Deng, Feng Wang, Feng Liu, Guangwei Ai, Guosheng Dong, Haizhou Zhao, Hang Xu, Haoze Sun, Hongda Zhang, Hui Liu, Jiaming Ji, Jian Xie, Juntao Dai, Kun Fang, Lei Su, Liang Song, Lifeng Liu, Liyun Ru, Luyao Ma, Mang Wang, Mickel Liu, MingAn Lin, Nuolan Nie, Peidong Guo, Ruiyang Sun, Tao Zhang, Tianpeng Li, Tianyu Li, Wei Cheng, Weipeng Chen, Xiangrong Zeng, Xiaochuan Wang, Xiaoxi Chen, Xin Men, Xin Yu, Xuehai Pan, Yanjun Shen, Yiding Wang, Yiyu Li, Youxin Jiang, Yuchen Gao, Yupeng Zhang, Zenan Zhou, and Zhiying Wu. 2023.
\newblock \href {https://doi.org/10.48550/ARXIV.2309.10305} {Baichuan 2: Open large-scale language models}.
\newblock \emph{CoRR}, abs/2309.10305.

\bibitem[{Zhang et~al.(2022)Zhang, Dellaferrera, Sikarwar, Armend{\'a}riz, Mudrik, Agrawal, Madan, Barbu, Yang, Kumar, Sadwani, Dellaferrera, Pizzochero, Pfister, and Kreiman}]{Zhang2022HumanOM}
Mengmi Zhang, Giorgia Dellaferrera, Ankur Sikarwar, Marcelo Armend{\'a}riz, Noga Mudrik, Prachi Agrawal, Spandan Madan, Andrei Barbu, Haochen Yang, T.~K.~Satish Kumar, Meghna Sadwani, Stella Dellaferrera, Michele Pizzochero, Hans-R{\"u}diger Pfister, and Gabriel Kreiman. 2022.
\newblock \href {https://api.semanticscholar.org/CorpusID:253801749} {Human or machine? turing tests for vision and language}.
\newblock \emph{ArXiv}, abs/2211.13087.

\bibitem[{Zheng et~al.(2023)Zheng, Chiang, Sheng, Zhuang, Wu, Zhuang, Lin, Li, Li, Xing, Zhang, Gonzalez, and Stoica}]{Zheng2023JudgingLW}
Lianmin Zheng, Wei-Lin Chiang, Ying Sheng, Siyuan Zhuang, Zhanghao Wu, Yonghao Zhuang, Zi~Lin, Zhuohan Li, Dacheng Li, Eric~P. Xing, Haotong Zhang, Joseph Gonzalez, and Ion Stoica. 2023.
\newblock \href {https://api.semanticscholar.org/CorpusID:259129398} {Judging llm-as-a-judge with mt-bench and chatbot arena}.
\newblock \emph{ArXiv}, abs/2306.05685.

\bibitem[{Zhou et~al.(2017)Zhou, Huang, Zhang, Zhu, and Liu}]{Zhou2017EmotionalCM}
Hao Zhou, Minlie Huang, Tianyang Zhang, Xiaoyan Zhu, and Bing-Qian Liu. 2017.
\newblock \href {https://api.semanticscholar.org/CorpusID:2024574} {Emotional chatting machine: Emotional conversation generation with internal and external memory}.
\newblock In \emph{AAAI Conference on Artificial Intelligence}.

\bibitem[{Zhou et~al.(2023)Zhou, Chen, Wan, Wen, Song, Yu, Huang, Peng, Yang, Xiao, Sabour, Zhang, Hou, Zhang, Dong, Tang, and Huang}]{Zhou2023CharacterGLMCC}
Jinfeng Zhou, Zhuang Chen, Dazhen Wan, Bosi Wen, Yi~Song, Jifan Yu, Yongkang Huang, Libiao Peng, Jiaming Yang, Xiyao Xiao, Sahand Sabour, Xiaohan Zhang, Wenjing Hou, Yijia Zhang, Yuxiao Dong, Jie Tang, and Minlie Huang. 2023.
\newblock \href {https://api.semanticscholar.org/CorpusID:265466372} {Characterglm: Customizing chinese conversational ai characters with large language models}.
\newblock \emph{ArXiv}, abs/2311.16832.

\end{thebibliography}
\bibliographystyle{acl_natbib}

% \newpage
\appendix

\section{Questionnaire Demonstration}

Each questionnaire consists of 11 questions: five are designed to collect anonymous personal information, five pertain to the Turing test, and one asks respondents to explain their strategies for distinguishing between human and AI during the test. For questions related to the Turing test, an example is provided in Table \ref{questionnaire}. The response of User \textit{B}, which is the focus of comparison, is highlighted in the questionnaire system for clarity.

\newcolumntype{L}[1]{>{\raggedleft\arraybackslash}p{#1}}

\begin{table*}[htbp]
    \centering
    \begin{tabular}{L{0.05\linewidth}p{0.85\linewidth}}
    \toprule
      \multicolumn{2}{l}{\textbf{Example Questionnaire}} \\
    \midrule
    \multicolumn{2}{l}{\textbf{Conversation 1}} \\
    1 & A: Been diving into Hesse's novels lately \\
    2 & \textcolor{teal}{B: Wow, Hesse's works are profound} \\
    3 & A: Steppenwolf \\
    4 & \textcolor{teal}{B: Is it Siddhartha or Steppenwolf?} \\
    5 & \textcolor{teal}{B: Oh} \\
    6 & \textcolor{teal}{B: His writing always provokes deep thought} \\
    7 & A: It's indeed a classic \\
    8 & A: Feels like a baptism lol \\
    9 & \textcolor{teal}{B: Exactly, like a journey for the soul, feels transformative after reading.} \\
    10 & A: His views really strike a chord with me, tbh \\
    11 & \textcolor{teal}{B: Yeah, his exploration of individualism and spiritual freedom hits right to the heart.} \\
    12 & A: I'm not much into philosophical thought, though \\
    13 & A: Just really enjoy works about the bourgeois \\
    14 & \textcolor{teal}{B: Then you might also like Kafka} \\
    15 & \textcolor{teal}{B: Another master at deeply dissecting human nature} \\
    16 & \textcolor{teal}{B: Especially with a sense of the times} \\

    \midrule
    \multicolumn{2}{l}{\textbf{Conversation 2}} \\
    17 & A: Been diving into Hesse's novels lately \\
    18 & A: Steppenwolf \\
    19 & \textcolor{teal}{B: Omg} \\
    20 & \textcolor{teal}{B: I just finished Demian} \\
    21 & A: It feels like a spiritual baptism \\
    22 & A: Wow, I love Demian too! \\
    23 & \textcolor{teal}{B: Because the Demian musical crew recently came to school} \\
    24 & \textcolor{teal}{B: Had to cram lol} \\
    25 & A: Yeah, literature and musicals seem to go well together \\
    26 & A: Are they here to perform \\
    27 & A: Or to do an exchange \\
    28 & \textcolor{teal}{B: It's a creative sharing session} \\
    29 & \textcolor{teal}{B: Sharing insights about their creation} \\
    30 & \textcolor{teal}{B: Some actors also performed parts of it} \\
    31 & A: Really? That's great \\
    32 & A: Your campus life is so rich, mate \\
    33 & \textcolor{teal}{B: I love musicals myself} \\
    34 & \textcolor{teal}{B: So it was a pleasant surprise hahaha} \\
    35 & \textcolor{teal}{B: And some of my favourite actors were at school!} \\

    \midrule
    \multicolumn{2}{l}{\textbf{Evaluation:}} \\
    & (A) User B in Conversation 1 is AI, User B in Conversation 2 is Human \\
    & (B) User B in Conversation 1 is Human, User B in Conversation 2 is AI \\
    \bottomrule
    \end{tabular}
    \caption{An example question from the questionnaire.}
    \label{questionnaire}
\end{table*}

\section{Prompt Demonstration}

\subsection{Chatbot Construction}
\label{sec: role prompt}

Table \ref{prompt0}-\ref{prompt1} present the prompts used to construct the chatbot under different dialogue settings. The dialogue history will be released after being accepted.

\begin{table*}[htbp]
    \centering
    \begin{tabular}{p{0.9\linewidth}}
    \toprule
    \textbf{Prompt: Ping-Pong Dialogue Chatbot} \\
    \midrule
    The below are social platform chat records from an undergraduate girl. You are not an AI but the specific person. "Response" refers to the message sent by the person. Determine the person's dialogue style, which includes tone, choice of vocabulary, punctuation marks, emoji, and typical topics of conversation. \\
Follow the person's established communication style to have new conversations with the user, incorporating similar expressions and emotions as used by the person:\\
1. You only need to reply in one short sentence like the utterances beginning with "Response", which can even be incomplete.\\
2. You don't need to use punctuation marks except for usage similar to the utterances. Do not use emojis in every response.\\
3. You should not always stick to the user. You can ask appropriate questions or start new related topics, but don't deviate from the current topic. \\
\\
\{Dialogue History\}\\
\\
User: Hi! How is it going?\\
Response:\\
    \bottomrule
    \end{tabular}
\caption{An example prompt for constructing a ping-pong dialogue chatbot.}
\label{prompt0}  
\end{table*}

\begin{table*}[ht]
    \centering
    \begin{tabular}{p{0.9\linewidth}}
    \toprule
    \textbf{Prompt: Burst Dialogue Chatbot} \\
    \midrule
    The below are social platform chat records from an undergraduate girl. You are not an AI but the specific person. "Response" refers to the message sent by the person. Determine the person's dialogue style, which includes tone, choice of vocabulary, punctuation marks, emoji, and typical topics of conversation.\\
Follow the person's established communication style to have new conversations with the user, incorporating similar expressions and emotions as used by the person:\\
1. You only need to reply in one short sentence like the utterances in the "Response", which can even be incomplete.\\
2. You don't need to use punctuation marks except for usage similar to the utterances. Do not use emojis in every response.\\
3. You should not always stick to the user. You can ask appropriate questions or start new related topics, but don't deviate from the current topic.\\
4. Both you and the user can send multiple consecutive messages. Consecutive messages are each labelled with sending timestamp and with "$\backslash$n" to concatenate, like [xxxx] ddddd$\backslash$n[xxxx] cccc$\backslash$n.... The timestamps of consecutive messages in the response should be incremental and at intervals that match the intervals shown in the example for humans replying to messages.\\
\\
\{Dialogue History\}\\
\\
User: [2024-06-10 10:34:22] Hi! \\
User: [2024-06-10 10:34:39] How is it going?\\
Response:\\
    \bottomrule
    \end{tabular}
\caption{An example prompt for constructing a burst dialogue chatbot.}
\label{prompt1}  
\end{table*}

\subsection{Topic Generation}

Before generating pseudo-dialogue, we generate 10 topics using GPT-4, Qwen-110B and Claude-3-Sonnet with the prompt "Generate 10 diverse topics for daily conversations without repetition.".

The topics generated by GPT-4 are:

\begin{itemize}
  \item \textbf{Travel Experiences}: Share your travel experiences and discuss future travel plans.
  \item \textbf{Culinary Exploration}: Talk about recent new dishes or restaurants you've tried, and recommend your favourites.
  \item \textbf{Movies}: Discuss the latest movies you've seen and share your impressions.
  \item \textbf{Book Recommendations}: Share the books you've recently read and discuss your reading list.
  \item \textbf{Health and Fitness}: Talk about your health habits and how to maintain vitality and wellness.
  \item \textbf{Technology Trends}: Exchange insights on emerging technologies, such as smart homes and AI technologies.
  \item \textbf{Learning Experiences}: Share insights from your course studies and discuss effective learning techniques.
  \item \textbf{Music and Arts}: Talk about your favorite music genres, musicians, or singers.
  \item \textbf{Cultural Differences}: Discuss the different cultures you have experienced and how to understand and adapt to these cultural differences.
  \item \textbf{Future Plans}: Discuss your short-term or long-term career goals and life plans.
\end{itemize}

The topics generated by Qwen-110B are:
\begin{itemize}
  \item \textbf{Weather Changes}: Discuss how to adjust your clothing or activity plans in current season.
  \item \textbf{Culinary Sharing}: Introduce special dishes or restaurants, discuss the taste of food, the environment, and unique dining experiences.
  \item \textbf{Movie Recommendations}: Evaluate the plot, performances of actors, or the director's style of recently watched movies.
  \item \textbf{Travel Stories}: Narrate interesting stories from your travels, experiences of cultural differences, or scenic spots, and explore the meaning and gains of traveling.
  \item \textbf{Pet Anecdotes}: Share the cute behaviors, habits of your pets, or warm interactions with them, reflecting deep pet-human relationships.
  \item \textbf{Work/Study Pressure}: Exchange methods for coping with stress, such as time management skills, relaxation activities, or psychological adjustment strategies.
  \item \textbf{Healthy Living}: Discuss the importance of a healthy diet and regular exercise, share wellness tips or personal fitness plans.
  \item \textbf{Musical Taste}: Explore how different music styles affect emotions and recommend newly discovered artists or songs.
  \item \textbf{Technology Insights}: Introduce the latest technological products and discuss how these innovations impact daily life or future prospects.
  \item \textbf{Weekend Plans}: Share your weekend plan and best ways to spend the leisure time.
\end{itemize}

The topics generated by Claude-3-Sonnet are:

\begin{itemize}
  \item \textbf{Recent Activities}: Updates on recent happenings and interesting events.
  \item \textbf{Entertainment}: Impressions and reviews of movies or TV shows watched recently.
  \item \textbf{Professional or Academic Insights}: Insights or achievements in studies or work.
  \item \textbf{New Hobbies or Plans}: Introduction to new hobbies or recently made plans.
  \item \textbf{Culinary Exploration}: Experiences with interesting foods and recommendations.
  \item \textbf{Travel Experiences}: Highlights from recent travels or trips.
  \item \textbf{Social Views}: Perspectives on recent societal events.
  \item \textbf{Interesting Encounters}: Encounters with interesting people or events.
  \item \textbf{Life Insights}: Recent insights or realizations about life.
   \item \textbf{Weather Impressions}: Thoughts and opinions on recent weather conditions.
\end{itemize}

\subsection{Pseudo-Dialogue Generation}

For each of the previously generated topics, we generate the corresponding pseudo-dialogue using the prompt in Table \ref{prompt3}-\ref{prompt4}.

\label{sec: pdg prompt}

\begin{table*}[htbp]
    \centering
    \begin{tabular}{p{0.9\linewidth}}
    \toprule
    \textbf{Prompt: Ping-pong Pseudo-Dialogue Generation} \\
    \midrule
    The below are social platform chat records from an undergraduate girl. You are not an AI but the specific person. "Response" refers to the message sent by the person. Determine the person's dialogue style, which includes tone, choice of vocabulary, punctuation mark, emoji, and typical topics of conversation. \\
Follow the person's established communication style to generate 10 consecutive rounds of ping-pong conversations between the user and response using the same format as the records, incorporating similar expressions and emotions as used by the person. \\
The conversation you need to generate is around the topic \{topic\}:\\
1. Both user and response only need to send message in one short sentence like the utterances, which can even be incomplete.\\
2. Both user and response don't need to use punctuation marks except for usage similar to the utterances. Do not use emojis in every response.\\
3. Both user and response should engage in natural chat. User should not keep asking questions. Response should not always agree with the user. Response can occasionally ask questions or introduce related new topics, but do not stray off topic. \\
4. Concatenate all the messages with $\backslash$n.\\
\\
\{Dialogue History\}\\
    \bottomrule
    \end{tabular}
\caption{An example prompt for generating ping-pong pseudo-dialogue with given topic.}
\label{prompt3}  
\end{table*}

\begin{table*}[htbp]
    \centering
    \begin{tabular}{p{0.9\linewidth}}
    \toprule
    \textbf{Prompt: Burst Pseudo-Dialogue Generation} \\
    \midrule
    The below are social platform chat records from an undergraduate girl. You are not an AI but the specific person. "Response" refers to the message sent by the person. Determine the person's dialogue style, which includes tone, choice of vocabulary, punctuation mark, emoji, and typical topics of conversation. \\
Follow the person's established communication style to generate 10 rounds of consecutive conversations (can send consecutive messages) between the user and response using the same format as the records, incorporating similar expressions and emotions as used by the person.  \\
The conversation you need to generate is around the topic \{topic\}:\\
1. Both user and response only need to send message in one short sentence like the utterances, which can even be incomplete.\\
2. Both user and response don't need to use punctuation marks except for usage similar to the utterances. Do not use emojis in every response.\\
3. Both user and response should engage in natural chat. User should not keep asking questions. Response should not always agree with the user. Response can occasionally ask questions or introduce related new topics, but do not stray off topic. \\
4. Both the user and response should send more than 3 multiple consecutive messages in each round. Consecutive messages are each labeled with sending timestamp and with "$\backslash$n" to concatenate, like [xxxx] ddddd$\backslash$n[xxxx] cccc$\backslash$n.... The timestamps of consecutive messages in the response should be incremental and at intervals that match the intervals shown in the example for humans replying to messages.\\
5. Concatenate all the messages with $\backslash$n.\\
\\
\{Dialogue History\}\\
    \bottomrule
    \end{tabular}
\caption{An example prompt for generating burst pseudo-dialogue with given topic.}
\label{prompt4}  
\end{table*}

\subsection{LLMs as Judge}
\label{sec: judge prompt}

We use the prompt in Table \ref{prompt5} to employ LLMs as Turing test judges. The option position is randomly assigned to avoid position bias.

\begin{table*}[h]
    \centering
    \begin{tabular}{p{0.9\linewidth}}
    \toprule
    \textbf{Prompt: AI Judge} \\
    \midrule
    You will be presented with two chat transcripts. In each transcript, User A is a human, while User B could be either a human or an AI. Based on the conversation content and style, your task is to determine in which chat transcript, the User B is a human or an AI. There are only two choices and the order does not indicate anything. You only need to print the option.\\
    Conversation 1: \{Conversation\_1\}\\
    Conversation 2: \{Conversation\_2\} \\
    \\
    (A) User B in Conversation 1 is AI, User B in Conversation 2 is Human\\
    (B) User B in Conversation 1 is Human, User B in Conversation 2 is AI\\
    \bottomrule
    \end{tabular}
\caption{An example prompt for constructing AI-based evaluation.}
\label{prompt5}  
\end{table*}

\end{document}